%% file: 0.main.tex
\documentclass[sigconf,article]{acmart}

\usepackage{enumitem}
\usepackage{bm}
\usepackage{multirow}

\usepackage{amsmath}
\usepackage{cases}

\usepackage{balance}

\usepackage[titletoc]{appendix}

\usepackage[]{fixltx2e}
\usepackage[bottom]{footmisc}
\usepackage{xcolor}
\usepackage{soul}
\usepackage{graphicx}
\usepackage{subcaption}
\usepackage{booktabs}
\usepackage[export]{adjustbox}

\AtBeginDocument{%
  \providecommand\BibTeX{{%
    \normalfont B\kern-0.5em{\scshape i\kern-0.25em b}\kern-0.8em\TeX}}}

\acmBooktitle{KDD '21: ACM SIGKDD Conference on Knowledge Discovery and Data Mining,
  August 14--18, 2021, Singapore}
\acmPrice{}
\copyrightyear{2024}
\acmYear{2024}
\setcopyright{rightsretained}
\acmConference[WWW '24]{Proceedings of the ACM Web Conference 2024}{May 13--17, 2024}{Singapore, Singapore}
\acmBooktitle{Proceedings of the ACM Web Conference 2024 (WWW '24), May 13--17, 2024, Singapore, Singapore}\acmDOI{10.1145/3589334.3648159}
\acmISBN{979-8-4007-0171-9/24/05}

\author{Nicholas Sukiennik}
\affiliation{
 \institution{Department of Electronic Engineering, Beijing National Research Center for Information Science and Technology\\Tsinghua University}
 \country{Beijing, China}}
\email{sukiennikn10@mails.tsinghua.edu.cn}

\author{Chen Gao$^*$}
\thanks{*Corresponding author.}
\affiliation{%
 \institution{Beijing National Research Center for Information Science and Technology, Tsinghua University}
 \country{Beijing, China}}
\email{chgao96@gmail.com}

\author{Nian Li}
\affiliation{
 \institution{Shenzhen International Graduate School, Tsinghua University}
 \country{Shenzhen, China}}
\email{linian21@mails.tsinghua.edu.cn}

\renewcommand{\authors}{Nicholas Sukiennik, Nian Li, Chen Gao}

\renewcommand{\shortauthors}{Nicholas Sukiennik, Nian Li, \& Chen Gao}

\begin{document}
\title[Uncovering the Deep Filter Bubble: Narrow Exposure in Short-Video Recommendation]{Uncovering the Deep Filter Bubble: Narrow Exposure in Short-Video Recommendation}

\begin{abstract} 
Filter bubbles have been studied extensively within the context of online content platforms due to their potential to cause undesirable outcomes such as user dissatisfaction or polarization. With the rise of short-video platforms, the filter bubble has been given extra attention because these platforms rely on an unprecedented use of the recommender system to provide relevant content. In our work, we investigate the deep filter bubble, which refers to the user being exposed to narrow content within their broad interests. We accomplish this using one-year interaction data from a top short-video platform in China, which includes hierarchical data with three levels of categories for each video. We formalize our definition of a "deep" filter bubble within this context, and then explore various correlations within the data: first understanding the evolution of the deep filter bubble over time, and later revealing some of the factors that give rise to this phenomenon, such as specific categories, user demographics, 
and feedback type. We observe that while the overall proportion of users in a filter bubble remains largely constant over time, the depth composition of their filter bubble changes. In addition, we find that some demographic groups that have a higher likelihood of seeing narrower content and implicit feedback signals can lead to less bubble formation.
Finally, we propose some ways in which recommender systems can be designed to reduce the risk of a user getting caught in a bubble.


\end{abstract}


\maketitle
\input{1.intro.tex}
\input{2.method.tex}

\input{3.experiments.tex}
\input{4.related.tex}
\input{5.conclusion.tex}

\section*{Acknowledgments}
This work is supported in part by the National Natural Science Foundation of China under 62272262, U23B2030, and 72342032, the National Key Research and Development Program of China under 2022YFB3104702, and the Guoqiang Institute, Tsinghua University under 2021GQG1005.

\balance

\bibliographystyle{ACM-Reference-Format}
\bibliography{citations}
\clearpage
\nobalance

\appendix

\end{document}

%% file: 1.intro.tex
\section{Introduction}\label{sec::intro}

Automatic recommendation has been the standard way to make sense of the unprecedented quantities of online information since their implementation on social media platforms like Facebook as early as 2010 \cite{jannach2010recommender}. While recommender systems have solved the problem of providing relevant content to users and creating a more user-friendly online experience, they have also given rise to some problems, namely, the filter bubble. The filter bubble was first described by Eli Pariser in 2011 \cite{pariser2011filter} as the phenomenon where the content seen on social media platforms becomes more and more personalized over time so that users eventually lack exposure to the full variety of content that exists.

More recently, the rise of short-video platforms has revolutionized both the way content is consumed and recommended on social media. Due to the single-column video layout ("Featured", or "For You Page"), users simply have to swipe up to enjoy an endless feed of unique content tailored to their interests. However, due to the lack of additional interest signals, the recommender must make use of largely implicit feedback on recommended content to learn users' preferences. Because on-platform interests are both learned and formed in the same space, the short-video recommender potentially has a larger risk of leading to the formation of a filter bubble. In other words, since there is no user-specified information such as search or browsing activity, the system has little information other than the feedback on existing videos to improve its ability to capture users' interests. This means it is possible that the recommender will only learn a small portion of user interests rather than all of them, and because most feedback is positive, it does not feel the need to learn more.

It is important to mention some concepts that are similar and often mentioned in the same context as the filter bubble: echo chamber and information cocoon. The concept of echo chamber is distinct from a filter bubble in that it refers to a scenario in which a user is embedded in a community where content of a certain standpoint or bias is continuously viewed \cite{nguyen_echo_2020} \cite{kitchens_understanding_2020}. The echo chamber could be seen as a potential negative outcome of the filter bubble, but the echo chamber typically refers to the entire set of information seen by a person, and not that coming from one individual platform \cite{cinelli2021echo} \cite{flaxman_filter_2016}. As such, we do not address the echo chamber explicitly in this work, leaving it to those in the fields of social science. 

Information cocoon, on the other hand, refers to a user's exposure to information on a specific platform that often covers similar topics, regardless of whether it was caused by user action or by the recommender system \cite{li_exploratory_2022}. In this way, we consider a filter bubble as a special case of information cocoon that is specifically related to the recommender system of a given platform. In this work, we address the filter bubble, as it is our primary interest to understand how the recommender system adapts to users' deeper interests.

Since its initial discovery, there have been many works addressing the filter bubble problem from several different perspectives, either to understand how it is formed or to alleviate it with novel recommendation models. As for the former, \cite{piao_humanai_2023} develops a model of the adaptive information dynamics between the users and the recommender algorithms, showing that user feedback is the essential factor in leading the recommender towards more or less diverse content. The presence of filter bubbles on short-video platforms was explored in \cite{li_exploratory_2022}. On the other hand, \cite{li_breaking_2023} design a simulation framework to model the recommender system pipeline of data collection, model learning, and item exposure in a loop, and propose a reinforcement learning (RL) method to select cross-community item exposures. Other works attempt to resolve the filter bubble via specialized model design \cite{gao_cirs_2023}, including those that are designed to increase diversity \cite{zheng_dgcn_2021}\cite{isufi_accuracy-diversity_2021}\cite{yang_dgrec_2023}. These models increase diversity while maintaining high accuracy. 
    \vspace{-.4em}
\subsection{User Decision Process}
User decision processes in online environments, influenced by behavioral science and recommender systems, form the basis for our investigation of the deep filter bubble. Central is the exploration vs. exploitation concept, indicating users either stick to familiar choices or explore new ones based on their desire to reduce uncertainty or maximize coherency, respectively \cite{riefer_coherency_2017}. Coherency maximization reflects a drive for choices consistent with past ones, and can significantly influence political preferences \cite{hornsby_how_2020}.

Studies on online music platforms reveal users' preferences evolve over time, with more diverse music consumption leading to higher engagement \cite{sanna_passino_where_2021, anderson_algorithmic_2020}. This diversity increase may represent online coherency maximizing exploration, contrasting with offline contexts where constant exploitation may lead to user boredom. Several other studies have emphasized the importance of diversity in recommendation \cite{kunaver2017diversity}\cite{lu2018diversity}\cite{kaminskas2016diversity}\cite{isufi_accuracy-diversity_2021}.

The mere exposure effect, where increased exposure to a stimulus enhances positive feelings towards it \cite{Zajonc_1968}, can impact recommender systems. Overemphasis on coherency can result in a filter bubble, with too much sameness leading to user boredom. Furthermore, the mere exposure effect could forge content-interests based on the recommender's offerings, potentially leading to less diverse content and a deep filter bubble. 

Understanding these behaviors, and the role of recommenders in shaping them, has social significance. Filter bubbles can undermine informed decision-making, essential in democratic societies \cite{Vasconcelos_Constantino_Dannenberg_Lumkowsky_Weber_Levin_2021} \cite{Santos_Lelkes_Levin_2021}. Moreover, online segregation, exacerbated by algorithmic bias and personalized feeds, can fuel inter-community tensions \cite{hosseinmardi_examining_2021}\cite{Haq_Tyson_Braud_Hui_2022}.

However, up until now, no works have investigated the hierarchical nature of the filter bubble. That is, rather than studying the formation of a bubble as subset of categories in relation to all the topics on the platform, no works have considered the narrowing of exposure to subcategories below those top-level interests. For example, a traditional filter bubble may mean that a user is seeing content mostly related to sports, rather than news, humor, pets, lifestyle, dance, etc. But a deep filter bubble would mean that within the topic of sports, a user's exposed content narrows over time from a variety of sports, to only soccer, and within soccer, only a given team. 

In the current work, we aim to address this gap by validating the presence of a "deep" filter bubble, namely the narrowing of exposure to subcategories over time. To do so, we leverage a dataset from one of the top short-video platforms in China, wherein each video is labeled with three-level hierarchical categories. We conduct an extensive exploration of the data to provide a novel understanding of the depth dimension of the filter bubble.  

   The contributions of this paper can be summarized as follows.
  \begin{itemize}[leftmargin=*]
      \item We develop a robust methodology for evaluating the presence and extent of the deep filter bubble.
      \item We conduct a comprehensive long-term analysis of the deep filter bubble and its correlating factors on a large real-world dataset.
      \item We provide insights as to how recommender models can prevent or alleviate the formation of the deep filter bubble and create more informative and gratifying content experiences for users. 

  \end{itemize}

The remaining part of this paper is organized as follows. We first formulate the methodology of our analysis in Section \ref{sec:method}, followed by the overall filter bubble analysis in Section \ref{sec::experiments} and the factors analysis in Section \ref{sec::factors}. Then, we present the literature review in Section \ref{sec::related} and discuss important implications in Section \ref{sec::discussion}. Finally, we conclude our paper in Section \ref{sec::conclusion}.

%% file: 2.method.tex

\section{Dataset and Methodology}\label{sec:method}
\begin{figure*}
    \centering
\adjincludegraphics[width=1.0\textwidth]{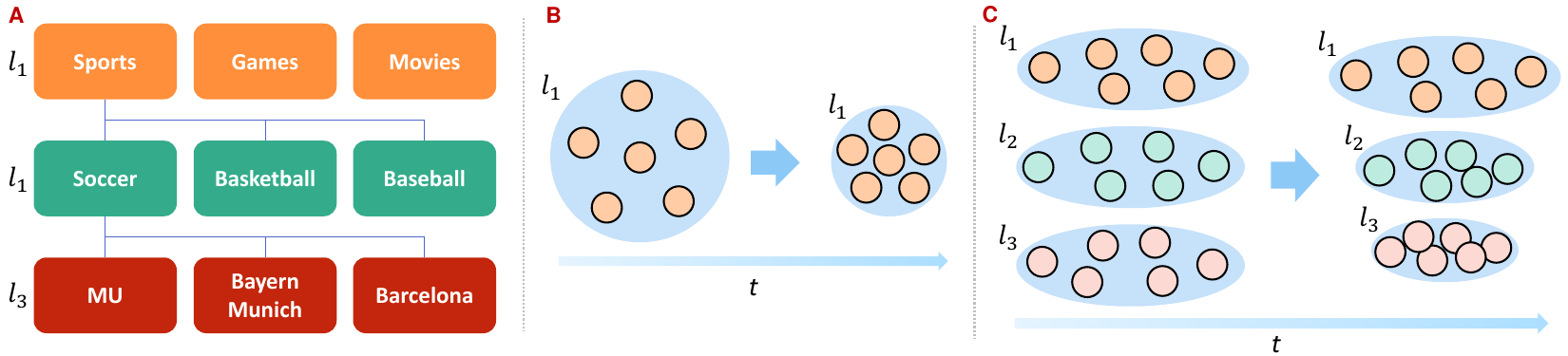}
    \vspace{-0.4cm}
    \caption{A conceptual overview of the traditional filter bubble, where A represents an example tree structure of hierarchical categories, and B and C represent the shallow and the deep filter bubbles, respectively, wherein the distance between items represents similarity. The items in B are assumed to be all uniform in level, whereas in C they are broken down into three levels. Figure C assumes that deeper-levels converge on more similar items over time.}
    \label{fig::concept}
\end{figure*}

\subsection{Dataset}
We use a large dataset of anonymized user-item interaction data from the single-column feed of a popular short-video application consisting of 400 million interactions, 400 thousand users, and 20 million items over a one-year period where all the users registered on the first day of the year. The data is filtered down to include only the videos that were viewed by the users and not those that were skipped immediately. The dataset includes three levels of hierarchical categories for each item and different forms of user feedback such as like, collect, and comment.


\subsection{Hierarchical Structure}

As our work heavily revolves around the hierarchical category structure of the data, we first motivate our investigation by presenting some basic statistics of this tree-like structure. 

\begin{figure}[h]
  \centering
  \begin{subfigure}{0.22\textwidth}
    \centering
    \includegraphics[width=1\textwidth]{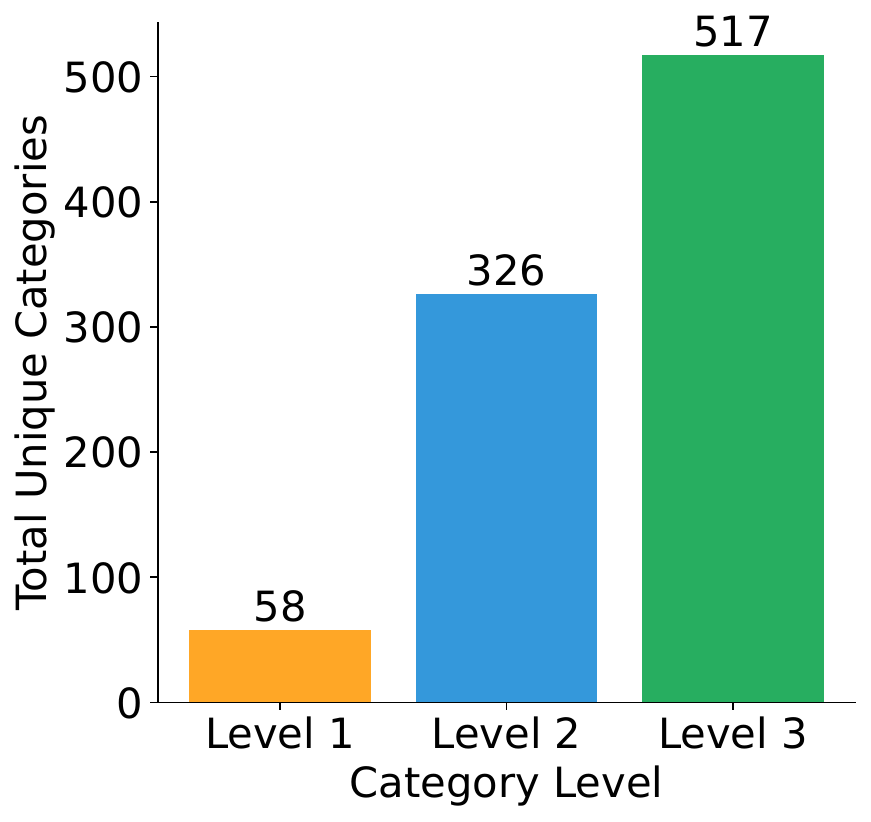}
    \caption{Categories per level \break}
    \label{fig:catelevel}
  \end{subfigure}
  \hfill
  \begin{subfigure}{0.22\textwidth}
    \centering
    \includegraphics[width=1\textwidth]{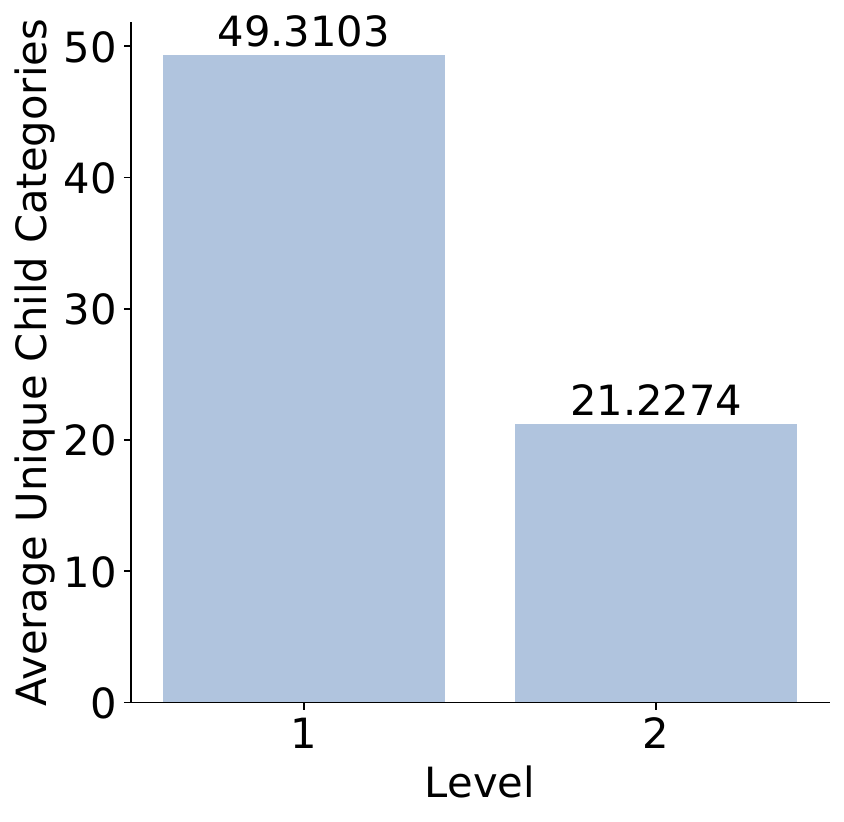}
    \caption{Average Children Per Item at Each Level}
    \label{fig:childrenlevel}
  \end{subfigure}
  \caption{Data Investigation}
  \label{fig:qualifyingdata}
\end{figure}

We first note the number of total unique categories per level across the whole dataset, depicted in figure \ref{fig:catelevel}: 58, 326, and 512 in level-ascending order. The number of categories increases with level, which is intuitive, as we would expect the tree to have more branches than roots. However, in figure \ref{fig:childrenlevel}, we note that the number of unique category children per level is, on average, higher for level 1 than for level 2. This is a counter-intuitive finding, as we would expect that there are more branches at the deeper levels considering that there are more deeper categories overall.

\subsection{Defining the Deep Filter Bubble}


Different works use different metrics to quantify the filter bubble depending on their data or the problem they are trying to solve. In our case, we are concerned with the decrease in exposure of deeper-level categories over time, so we must devise a metric that accurately reflects this phenomenon. To quantify this, we use a normalized form of coverage, where coverage is defined as the number of unique categories seen by a user over a specified time period. 

We formally define the metric for the filter bubble at each level by taking the coverage as a ratio of categories seen versus the total possible categories seen for a given level.

\begin{equation}\label{eqn:naivecoverage1}
\mathcal{C}_l = \frac{{n_{\text{seen},l}}}{{n_{\text{total},l}}}
,\end{equation}
where $\mathcal{C}_l$ is the coverage at a given level,
$n_{\text{seen,l}}$ is the number of categories seen at that level, and
$n_{\text{total,l}}$ is the total number of possible categories at that level.

We can also compute an overall coverage by summing the seen categories and total categories over all the levels:

\begin{equation}\label{eqn:naivecoverage2}
\mathcal{C} = \sum_{l=1}^{N} \frac{{n_{\text{seen},l}}}{{n_{\text{total},l}}}. 
\end{equation}

These equations allow us to quantity the conceptual phenomenon of the deep filter bubble, as illustrated in Figure \ref{fig::concept}, using the coverage of categories to represent the similarity of items exposed to each user.

%% file: 3.experiments.tex
\section{Overall Deep Filter Bubble Characterization}\label{sec::experiments}

\subsection{Diversity Investigation}
\begin{figure}[h]
  \centering
  \begin{subfigure}{0.23\textwidth}
    \centering
    \includegraphics[width=1\textwidth]{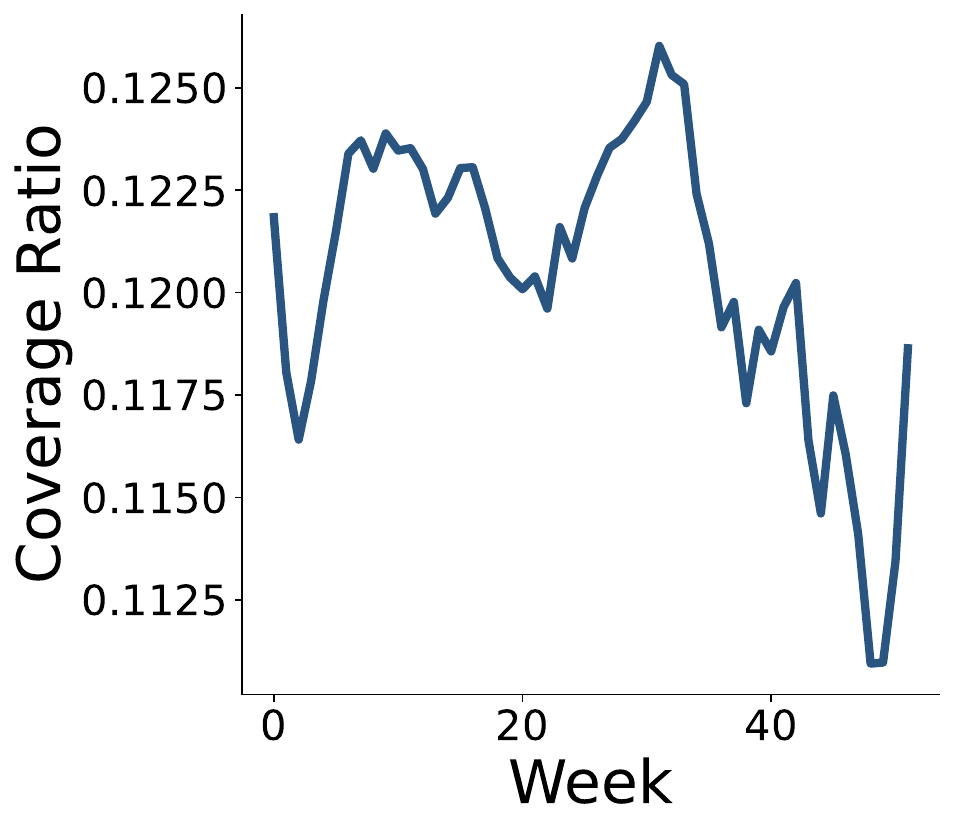}
    \caption{Overall coverage over time.}
    \label{fig:overall_coverage}
  \end{subfigure}
  \hfill
  \begin{subfigure}{0.22\textwidth}
    \centering
    \includegraphics[width=1\textwidth]{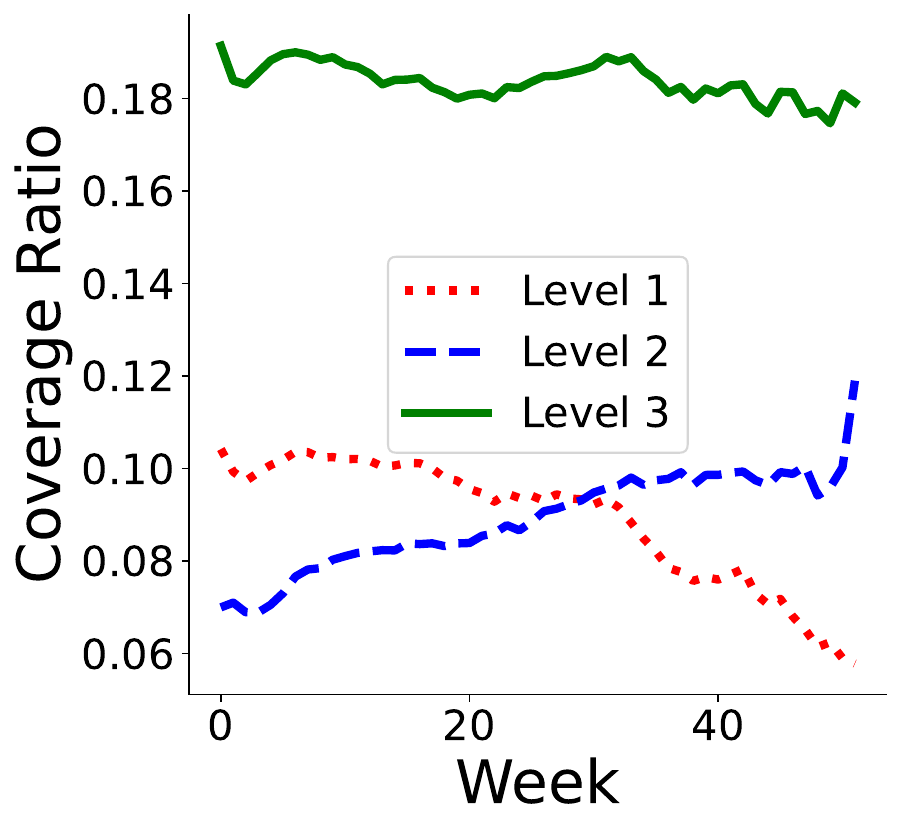}
    \caption{Coverage over time per level.}
    \label{fig:overallcoverageperlevel}
  \end{subfigure}

  \caption{Coverage over time.}
  \vspace{-.2cm}
  \label{fig:traditionalfb}
\end{figure}

In order to present results in an easily interpretable format, we split our interaction data into 52 time windows representing the weeks of the year, where user interaction statistics are consolidated into weekly statistics by taking their averages. 
Using the definition of low diversity as being a strong indicator of filter bubble formation, we conduct a preliminary analysis of the diversity, in terms of coverage ratio, for all users. This metric is provided on an overall and per-level basis in Figure \ref{fig:traditionalfb}.

 From the overall coverage in Figure \ref{fig:traditionalfb}a, we can see an obvious decrease in the overall coverage ratio towards the end of the year, despite some fluctuations throughout. In Figure \ref{fig:traditionalfb}b, we can start to determine the reasons for the trend and fluctuations. Qualitatively, we can state that the coverage decrease is caused primarily by a decrease in top-level category exposure, whereas the second level counteracts that decrease to some extent, and also causes a spike at the end. At the same time, the third level is mostly constant over time, with some rapid fluctuations towards the end. From this perspective, we can define three distinct possible filter bubbles: one for each category level. Based on the quantitative analysis, we can conclude that there is an obvious filter bubble being formed on the first level, no filter bubble formation on the second level, and minimal filter bubble formation on the third level. 
 
 

\subsection{Evolution over Time}\label{sec:timebubble}
In a platform of millions of users, it is possible for the recommender to respond differently according to different users' needs and interests. With this in mind, we design a criterion to label a user as "in" or "out" of a filter bubble at each time window. To do this, we specify a coverage threshold: if the categories seen by a user are less than the median number of categories seen by all users on a given level, then we consider that user to be "in" the filter bubble at that level:

\begin{equation}\label{eqn:criterion}
  \mathcal{B}_{u,t} = 
  \begin{cases}
    \text{"in"}, & \text{if } C_{u,t} < \text{Median}(C_{\cdot,t}) \\
    \text{"out"}, & \text{otherwise}
  \end{cases},
\end{equation}

\noindent where $\mathcal{B}_{u,t}$ is the status of user u at time window t, indicating whether they are "in" or "out" of the filter bubble, 
$C_{u,t}$ is the number of categories seen by user u during time window t and 
$\text{Median}(C_{\cdot,t})$ is the median number of categories seen by all users during time window t. On this basis, we proceed to calculate the proportion of users who are in the filter bubble at each level for each week in the year.


\begin{figure*}
    \centering
\adjincludegraphics[width=.99\textwidth
]{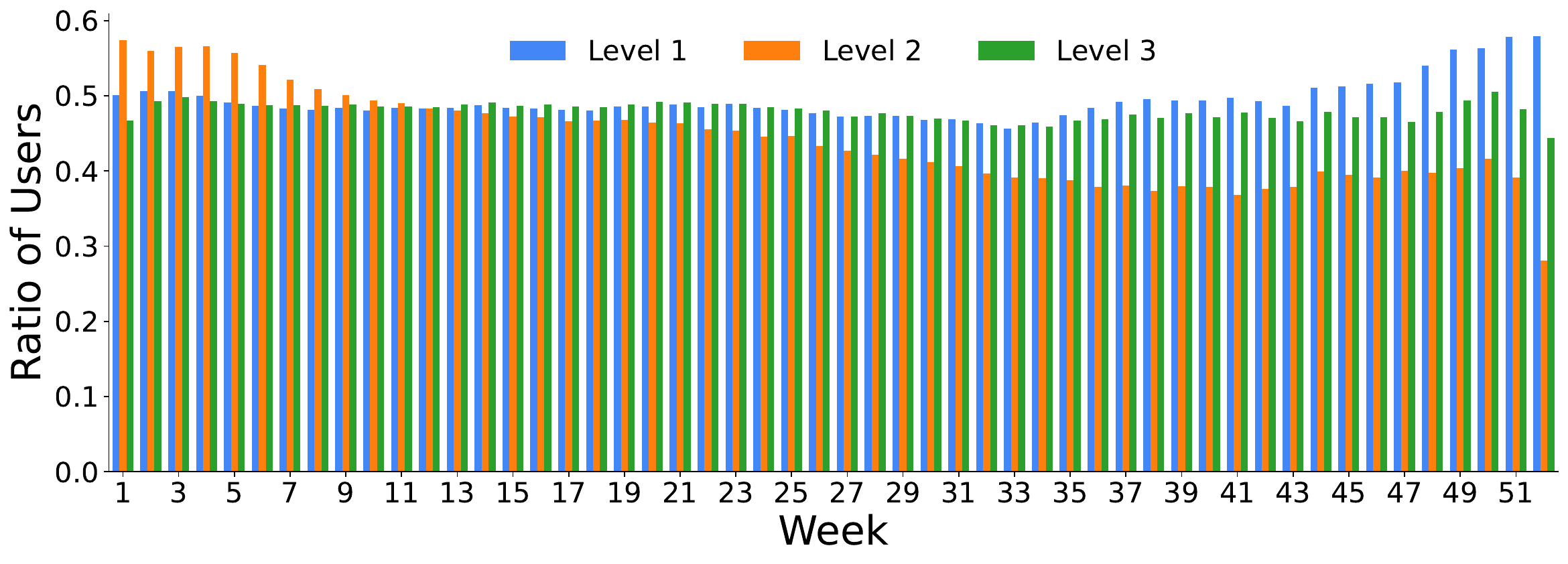}
    \vspace{-0.4cm}
    \caption{Ratio of users in bubble for each level over time.}
    \label{fig::ratiosovertime}
\end{figure*}

From Figure \ref{fig::ratiosovertime}, we can observe that as time progresses, the proportion of users in the filter bubble at each level varies in unique ways. For level 1, while remaining constant for most of the year, the number of users in the filter bubble increases within the final 20 weeks of the year. Level two, on the other hand, dominates the deep filter bubble ratio at the beginning, whereas by the end of the year, it significantly decreases, having the lowest filter bubble ratio. Over the same period, the proportion of users in the bubble for level two decreases significantly and then pans out by the end of the year before a final drop. Interestingly, the filter bubble for the deepest level remains relatively constant most of the year, even decreasing slightly before fluctuating during the final 20 weeks of the week. Both levels 2 and 3 have a substantial drop in the last three weeks of the year, resulting in an end state where where filter bubble of level 1 is dominant, followed by levels 3 and 2 respectively. 

The conclusions we can gather here are that upon first registering for the app, a user is more likely to see narrower content at a medium depth level, but over time, the content of the medium and deep levels grow more diverse, and the content in the top level gets narrowest. To use an intuitive example, if we consider a just-registered user who enjoys watching videos about games, movies, and music, the videos that they watch within each of them will span a variety of genres: for games, puzzle, sports, and strategy games; for movies, action, comedy, and romance movies; and for music, rock, electronic, and pop. But by the end of the year, the recommender will learn that they prefer games more than movies and music, but within games, they will now see four or five different game genres. Meanwhile, the specific games, film actors/directors, and music artists under each genre will remain equally diverse over time. Assuming that the recommender is accurately learning users' real interests, we can conclude that a user who likes a specific form of media would be more interested in various genres within that form of media with more likelihood than they would be interested in two entirely distinct forms of media. Expanding to the general scenario, we can state that diversity of second-level categories is more important than diversity of top-level categories \textit{once} the users' top-level interests are well understood, but not before. 

However, if we consider that a filter bubble can give rise to an information cocoon or echo chamber, we should also pay attention to whether the narrowing of top-level categories is truly beneficial to the user over the long term and consider the trade-off between platform engagement and user satisfaction. 

\vspace{-0.5em}
\subsection{Distribution over Users}

\begin{figure}[h]
  \centering
  \begin{subfigure}{0.23\textwidth}
    \vspace{1em}
    \centering
    \includegraphics[width=1\textwidth]{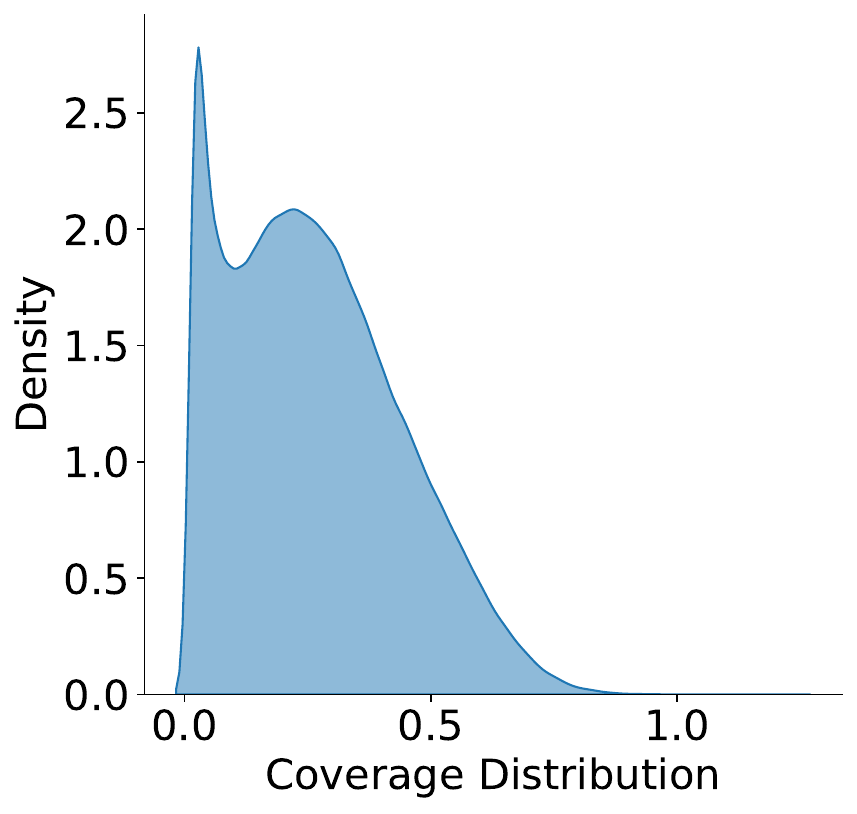}
    \caption{Distribution of User Coverage over all Levels}
    \label{fig:overall_distr}
  \end{subfigure}
  \hfill
  \begin{subfigure}{0.22\textwidth}
    \centering
    \includegraphics[width=1\textwidth]{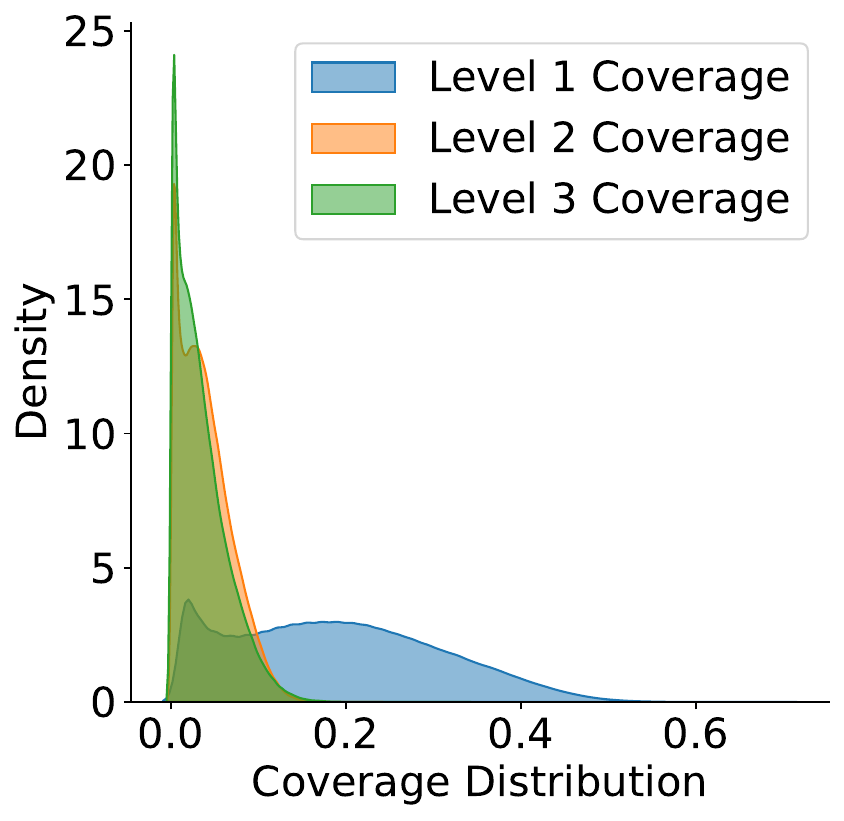}
    \caption{Distribution of User Coverage  Per Level}
    \label{fig:distrperlevel}
  \end{subfigure}
      \vspace{0em}
  \caption{Coverage probability density distribution overall and for each level.}
  \label{fig:distributions}

\end{figure}

Because our metric for filter bubble presence is relative to the number of categories seen by other users, we should also define an absolute metric to determine the tendency for category-levels themselves to have a stronger tendency to form filter bubbles, regardless of the relative behaviors of the users viewing those category-levels. To do so, we normalize the coverages per user using equation \ref{eqn:naivecoverage1}, and then we calculate the probability density of each coverage ratio. Figure \ref{fig:distributions} depicts the resulting probability density distributions for both the overall coverage and the coverage per level. 
By decomposing the overall distribution into a separate distribution for each category level, we can understand the reasons for the contours of the probability density. In Figure \ref{fig:overall_distr}, there is a spike in probability density at a very low coverage ratio, and there is also a seemingly separate Gaussian distribution centered around approximately 0.35. By looking at Figure \ref{fig:distrperlevel}, we can see that the initial spike is due to the fact that the probability densities for both levels 2 and 3 are extremely sharp at a low value between 0 and 0.1. Meanwhile, although there is a small initial spike for level 1, the distribution mostly centers in a normal manner around 0.2.

From this, we can conclude that, although the relative filter bubble was the strongest for level 1, as determined by the results of section \ref{sec:timebubble}, it also turns out that level 1 is the level with the most equitable distribution of coverages over all users. On the other hand, it turns out to be levels 2 and 3 that display a distribution of coverages that are restricted to a narrow low range. This result could be partially explained by the results in Figure \ref{fig:childrenlevel}, which we recall reveals that level 1 categories, on average, have more children than their level 2 counterparts. At the same time, each level-category's children only make up a fraction of the categories for the following level. This tree structure means that a user who only sees a few upper-level categories has a specific limited number of deeper categories that are \textit{possible} to see, given that the other categories are children of parents whom they also did not see. In this way, it is intuitive that our distribution of coverage will decrease the deeper the category level. We can generalize this phenomenon by saying that a user who has seen a certain fraction of categories at an upper level will necessarily see a smaller absolute fraction of categories at the lower levels, assuming that children are evenly dispersed.  

From a user standpoint, it is natural that users' interests would be more narrow as they get deeper. But from a platform perspective, it is important to broaden upper-level category exposure so that users do not get stuck in deeper-level absolute bubbles based on their incomplete upper-level exposure.  

Now that we have an understanding of the overall state of the deep filter bubble for our dataset, we proceed with an analysis of the factors that could potentially contribute to the formation of the deep filter bubble from three angles: 
specific categories, user characteristics, and feedback. 



\section{Factors Analysis}\label{sec::factors}

\subsection{The Effect of Specific Categories}

To determine whether certain categories have a "pulling" effect, i.e. cause the filter bubble to be pulled into the filter bubble more so than other categories, we perform an analysis of the probabilities with which users switch from category to category over time. Since each user sees potentially hundreds of videos per week, we must devise a concise way of classifying the important categories for each user for which to compute the probabilities. To do so, we select the 50 dominant categories per user for each level. Then we calculate the probability of each of these dominant categories changing to another category from week to week. Finally, we average the probabilities over all users and all weeks, providing us with a general picture of the pulling effect of the different categories for each level. 

\begin{figure}[h]
  \centering
  \begin{subfigure}{0.22\textwidth}
    \centering
    \includegraphics[width=\linewidth]{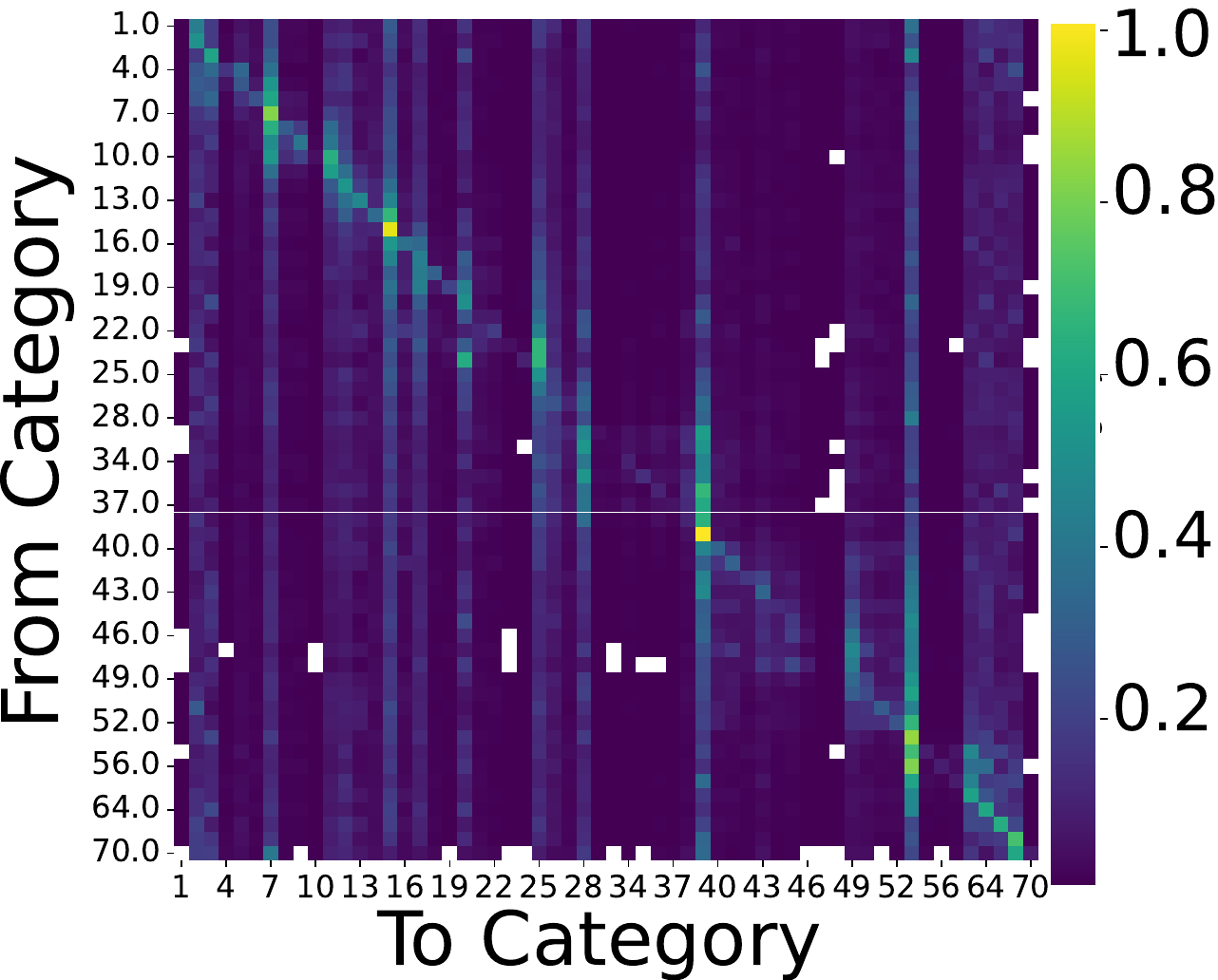}
    \caption{Level 1}
    \label{fig:probs_1}
  \end{subfigure}
  \hfill
  \begin{subfigure}{0.22\textwidth}
    \centering
    \includegraphics[width=\linewidth]{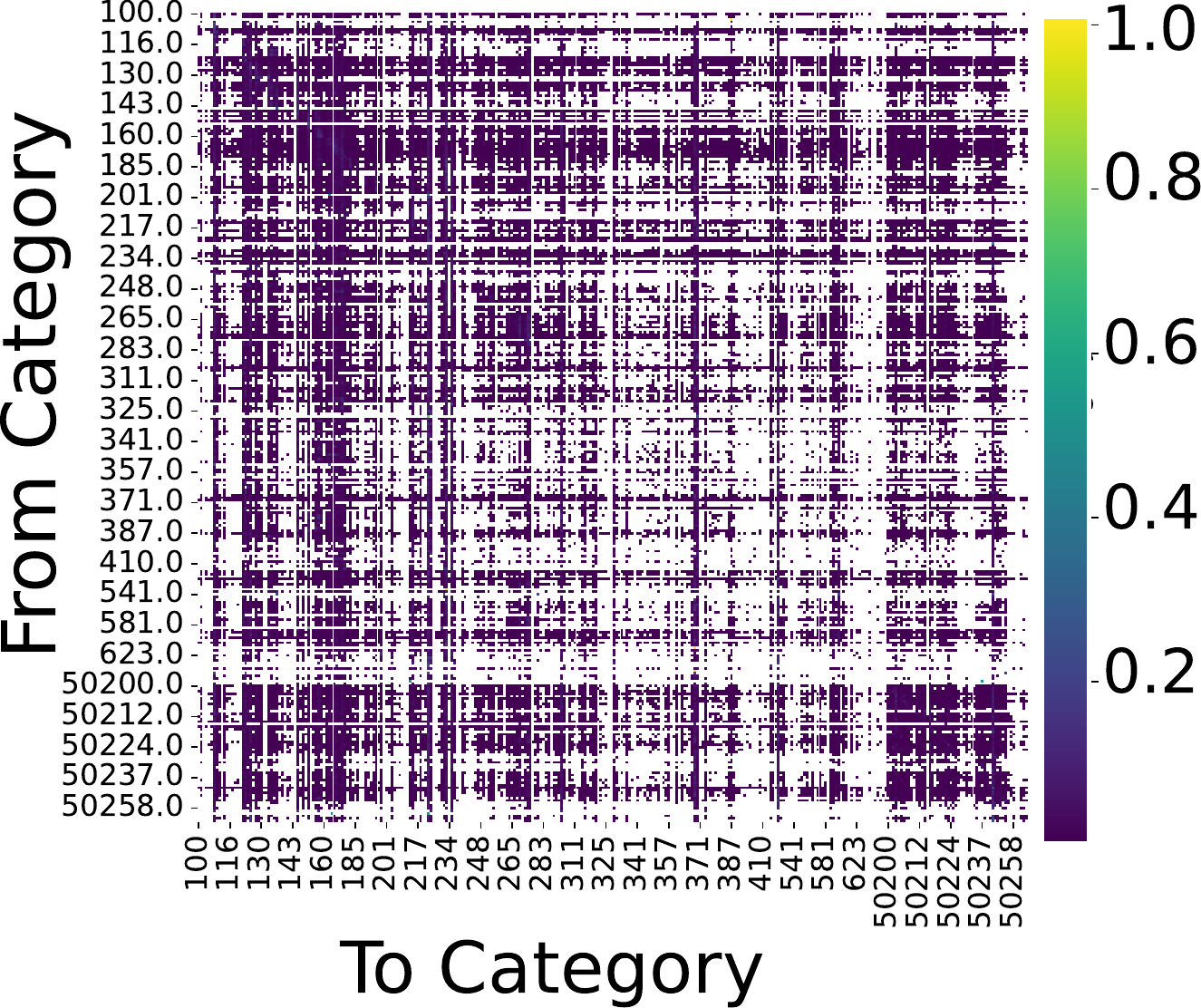}
    \caption{Level 2}
    \label{fig:probs_2}
  \end{subfigure}
  \vspace{0.5cm}
  \begin{subfigure}{0.22\textwidth}
    \centering
    \includegraphics[width=\linewidth]{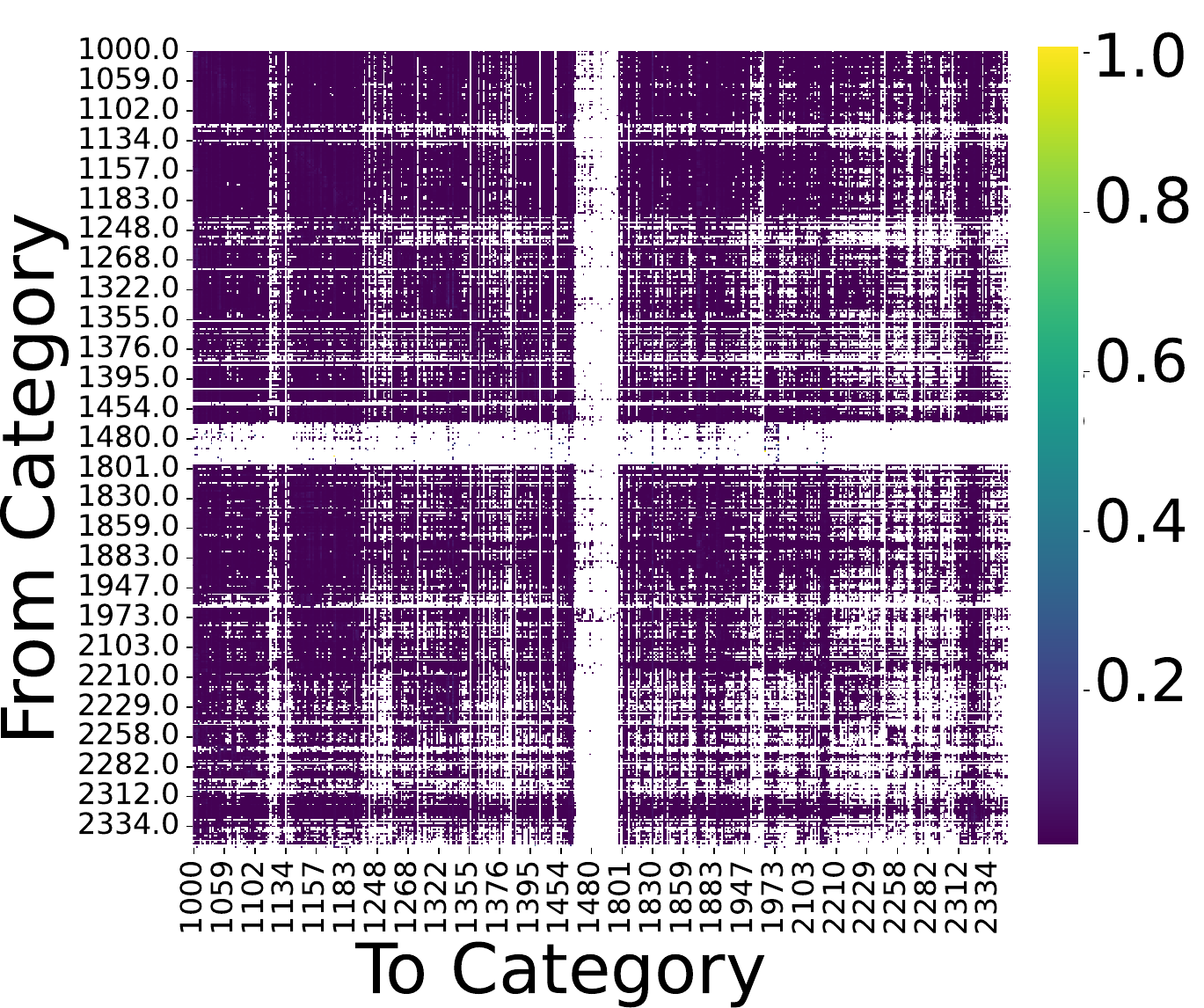}
    \caption{Level 3}
    \label{fig:probs_3}
  \end{subfigure}
    \vspace{-.2cm}
  \caption{Probability of switching dominant categories at any point in time for each level.}
  \vspace{-.1cm}\label{fig:probs_3levels}
  \end{figure}

The results, in Figure \ref{fig:probs_3levels}, tell us that for level 1, users are more likely to remain viewing the same categories from week to week, which is in line with our expectations. We would assume that a user has a stronger opinion on top-level, broad, categories, than deeper sub-categories. This tells us that for the top level, there is a potential of being pulled into a filter bubble based on initial interests.
In Figure \ref{fig:probs_2}, we note that there is no specific pattern of transition, telling us that a user may have a wide variety of medium-depth interests and that none of them has a particular pulling effect. A similar conclusion can be drawn from Figure \ref{fig:probs_3} for the third-level categories. 
From this analysis, we conclude that there are essentially no specific categories that have an overwhelming pulling effect on users' viewing activity. However, we also conclude that for top-level interest, there is a chance of a filter bubble forming based on the categories viewed initially. 
  
\subsection{Effects of User Characteristics}
\begin{figure*}[h]
    \centering
\adjincludegraphics[width=0.99\textwidth
]{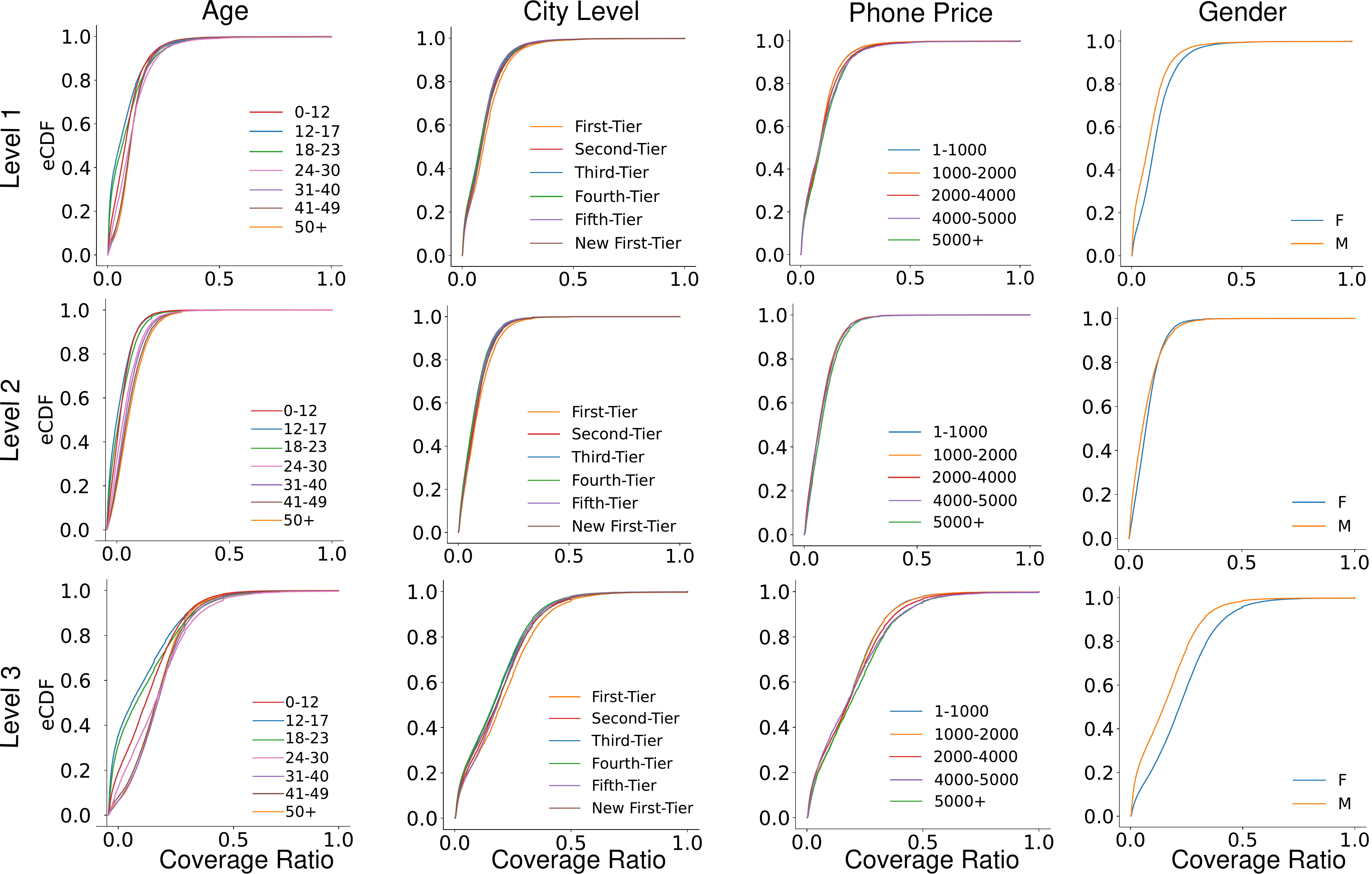}
    \vspace{-0cm}
    \caption{Empirical cumulative distribution function (eCDF) of average coverage ratio over all users at each level for four different demographic factors.}
    \label{fig::demog}
\end{figure*}
We conduct an extensive analysis of user characteristics with respect to category coverage at each level in Figure \ref{fig::demog}, with the goal of determining if there are any personal factors that could lead to a user being more or less prone to get trapped in a filter bubble. To do so, we plot an empirical cumulative distribution function to depict the strength of the relationship between category coverage and each of the four demographic characteristics: age, city level, phone price, and gender. The first thing we note is that for all demographics and levels, there is virtually no variation in probability for any of the variables after a coverage of 0.5. This is because the vast majority of users' coverages are in the lower half of the coverage range. Thus, we will focus our analysis on the first half of the coverage ranges to determine if there are any significant correlations between users' characteristics and coverage. 

\subsubsection{Age}
First, we look at the correlations between age and coverage ratio. Looking closely, we notice that there is a slight relationship between age and coverage, the one in level 3 being the most salient. The trend shows us that younger users have a higher probability of having a lower coverage ratio. This could pose a risk for younger users to get stuck in a deep filter bubble more easily than older users. 

\subsubsection{City Level}

Since our data was collected in China, we classify the cities of each anonymized user by their city-tiers, a classification that roughly corresponds to the size and economic strength of a city in China.
Although the relationship between city-tier and coverage is not especially strong, the figures for all three levels do show that users in first-tier cities have the highest relative coverage compared to users in other cities, and slightly more so in level 3, suggesting that users in lower tier-cities could be more likely to get stuck in a deep filter bubble. 

\subsubsection{Phone Price.}

Phone price has been used in prior works as a proxy metric for income \cite{li_exploratory_2022}. In our dataset, there is no exceedingly evident trend between phone price and coverage. 

\subsubsection{Gender.}
From the figures, we can gather that gender indeed plays a part in users' exposure to diverse categories, but only on levels 1 and 3. For these two levels, there is a tendency for males to have lower coverage than females, implying that males could be more likely to get caught in a shallow or deep filter bubble.
    \vspace{-0.4em}
\subsection{Influence of Feedback}

A user's feedback represents their level of satisfaction or dissatisfaction with the content provided by the recommender on a social media platform. However, different types of feedback can have different meanings. Namely, implicit and explicit feedback are used in different ways by the recommender. Implicit feedback is much more abundant than explicit feedback and is therefore much more commonly used to train recommendation models. However, there are still questions as to how to use different forms of feedback to optimally learn user preferences \cite{Pan_Gao_Chang_Niu_Song_Gai_Jin_Li_2023}. For example, if a user finishes watching a video but doesn't tap the like button, does she like that video? Or does "liking" override the fact that a video was not watched entirely through?

In this section, we will consider the ways in which different forms of feedback inform the recommender in different ways, and investigate in which ways if any, these different forms of feedback lead to different trends in the formation of the filter bubble. If there is a difference in trend between implicit and explicit feedback in the formation of the filter bubble, this could be seen as the recommender making inferences about which signals have more or less power to explain user interests. As such, by conducting the analysis of feedback type with respect to filter bubble formation, we aim to determine whether the recommender itself could be in part responsible for the creation of the filter bubble. 

   \begin{figure}
  \centering
  \begin{subfigure}{0.48\textwidth}
    \centering
    \includegraphics[width=\linewidth]{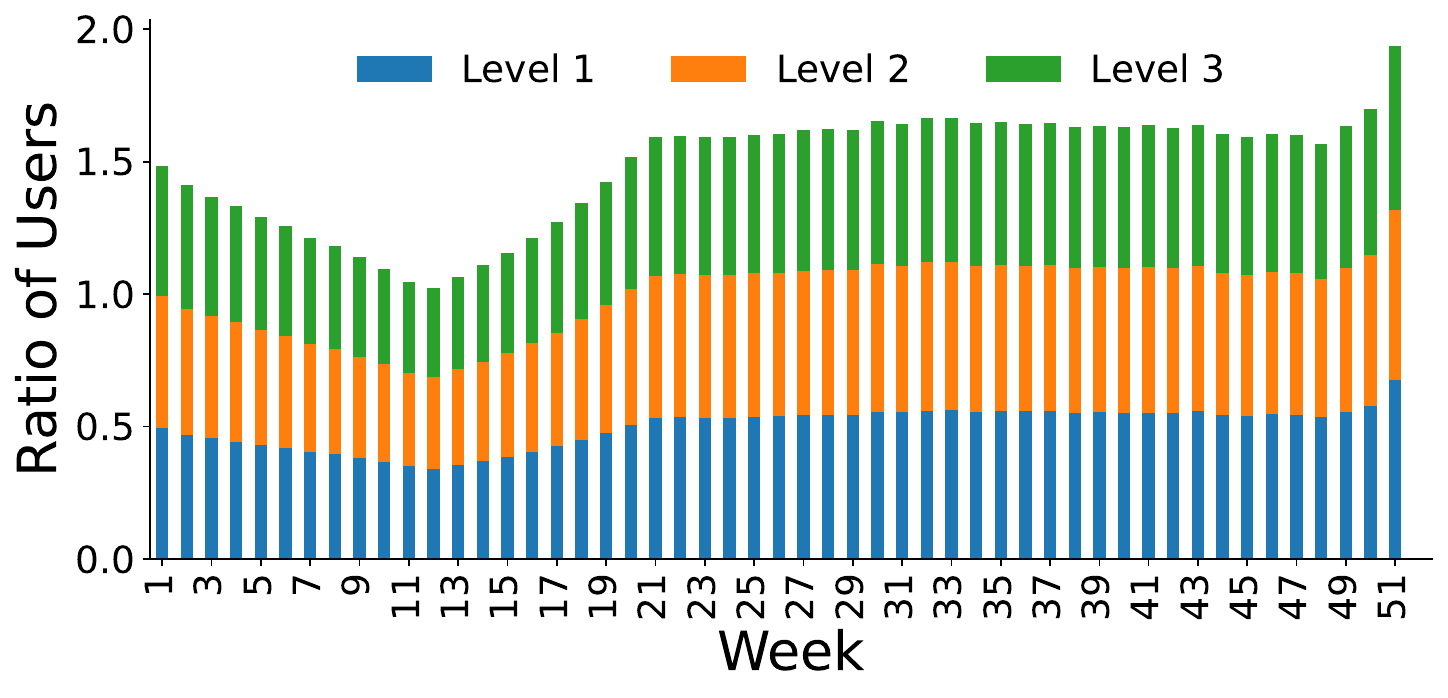}
    \caption{Explicit feedback.}
    \label{fig:feedback_bubble_expl}
  \end{subfigure}
  \begin{subfigure}{0.48\textwidth}
    \centering
    \includegraphics[width=\linewidth]{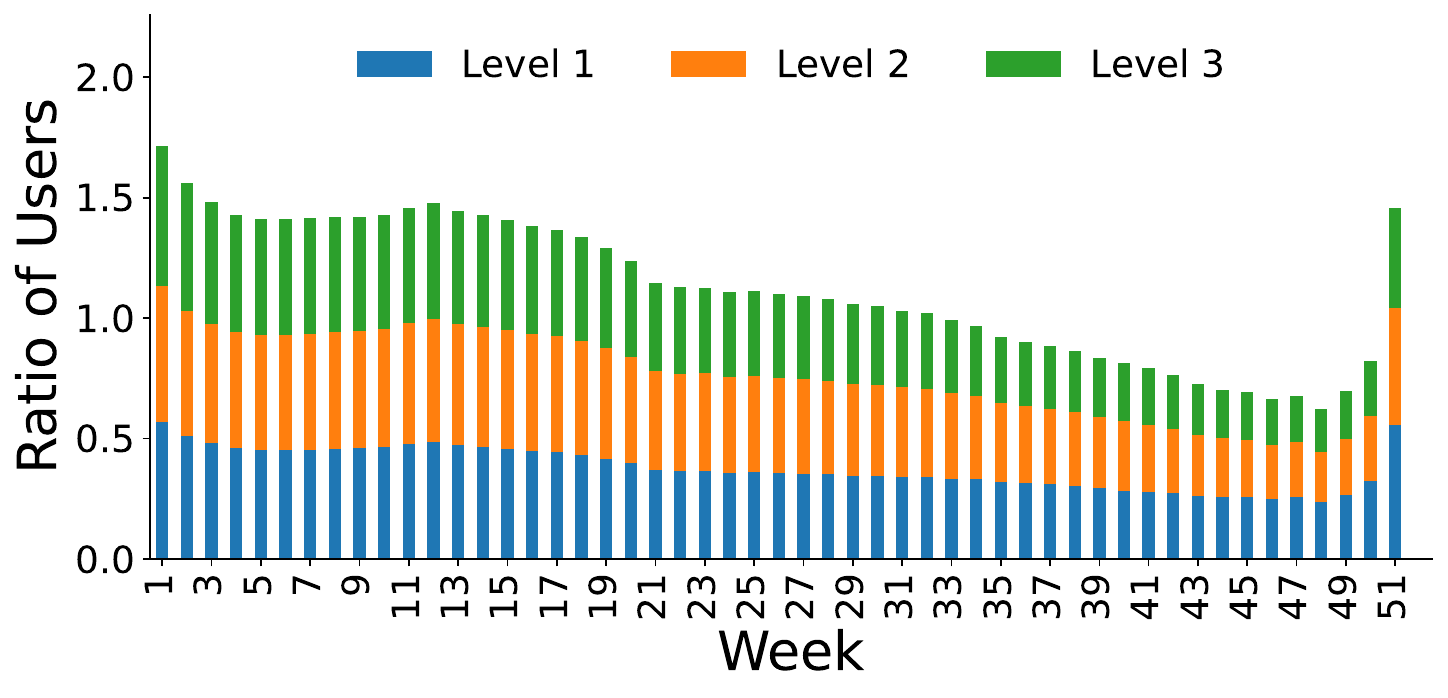}
    \caption{Implicit feedback.}
    \label{fig:feedback_bubble_impl}
  \end{subfigure}
  \caption{Progression of three-level filter development over time using only interactions with explicit and implicit feedback, respectively. }
\vspace{-1.5em}
  \label{fig:feedback_bubble}
\end{figure}

To conduct our analysis, we first filter down interactions for all users based on what form of feedback they gave to the viewed item. We consider implicit and explicit forms of positive feedback. For implicit feedback, we make use of the "click" metric, which, contrary to its name, simply means that the user has watched a video for more than half of its duration. On the other hand, for explicit feedback we consider "like", "follow", "share", and "comment". Interactions are preserved in the explicit data subset if they include one or more of the explicit forms of feedback. For implicit interactions, we specifically filter out those that have any of the explicit forms of feedback, whereas explicit does not necessarily imply a non-click.

Next, we recalculate the coverage at each time window for each user and then use the criterion defined in equation \ref{eqn:criterion} to label users as "in" or "out" of the filter bubble at each time window. In Figure \ref{fig:feedback_bubble}, we plot the resulting ratios of users in the bubble for each level. The ratios for all three levels are stacked on top of each other for space considerations. 

Let us first assess the results for Figure \ref{fig:feedback_bubble_expl}, explicit feedback. Immediately we notice that there is a drastic and nearly even decrease in filter bubble for all levels in the first three months of the year. Afterward, the ratio for all levels once again goes up, and within 10 weeks surpasses where it was initially. From then on, the ratios remain practically constant until the very end of the year, at which point they increase, being lifted up primarily by the increase in level 1. Overall, this is a very different trend from the overall trend we saw in Figure \ref{fig::ratiosovertime}, aside from the increase of level 1 at the end. This tells us that the explicit feedback does indeed pose a relationship with the formation of the deep filter bubble. In order to define the nature of that relationship, we shall proceed to compare this trend with the one for implicit feedback as seen in Figure \ref{fig:feedback_bubble_impl}. 

The figure relating the progression of filter bubble across three layers over time using only the subset of interactions with implicit feedback is shown in Figure \ref{fig:feedback_bubble_impl}. This figure shows yet another trend, distinct from both explicit feedback interactions in Figure \ref{fig:feedback_bubble_expl} and the overall interactions in Figure \ref{fig::ratiosovertime}. This figure shows an almost consistent downward trend in filter bubble ratio for all three category-levels from the start to week 48, at which point the filter bubble ratio for all three levels jumps. Indeed, this final jump is the only similarity in trend across all three figures. This spike could be a result of user sampling bias due to the fact that all the users registered at the beginning of the year.

Most important for our current analysis, however, is the comparison between the trends in implicit and explicit feedback-induced deep filter bubbles. What we can conclude is that interactions with implicit feedback can lead to a decrease in the filter bubble over time in comparison with those explicit feedback interactions. This implies that the recommender is potentially leveraging explicit feedback too strongly in its modeling of the users' interests over time. While explicit feedback can clearly represent stronger preferences than implicit feedback, it is important for the recommender to moderate the impact of strong preferences so that they do not dominate the exposed content and push out items from more diverse categories. The overemphasis on modeled interests represents one of the ways that the recommender itself could be responsible for the formation of the filter bubble. 

%% file: 4.related.tex
\section{Related Work}\label{sec::related}
\subsection{Filter Bubble}
The concept of the filter bubble has been growing in prominence since it was first proposed in Eli Pariser's 2011 book \cite{pariser2011filter}, detailing how over-personalization can lead to intellectual isolation. A subsequent study by Nguyen \textit{et al.} \cite{nguyen_exploring_2014} examines the temporal impacts of a collaborative filtering recommender on the diversity of items to better quantify the filter bubble effect. More recently, McKay \textit{et al.} \cite{mckay2022turn} attempts to shine light on the mechanisms of filter bubble from a social perspective by conducting studies on users' needs for diverse perspectives. 
Meanwhile, several works approach the filter bubble from a technical perspective, designing recommender models specifically to alleviate the problem. Li \textit{et al.} \cite{li_breaking_2023} proposes a reinforcement learning framework for controllable recommendation to alleviate filter bubbles by adaptively selecting diverse item connections, whereas Gao \textit{et al.} \cite{gao_cirs_2023} proposes an interactive recommender system using reinforcement learning and causal inference to model item overexposure effects on user satisfaction.
Some works attempt to explain the underlying mechanisms of filter bubbles. Li \textit{et al.} \cite{li_exploratory_2022} is the first one to conduct an exploratory study of the filter bubble created by short-video platforms, while \cite{piao_humanai_2023} creates a structural model to explain their formation as a result of human and AI interactions. 
Several works examine the potentially harmful societal effects of filter bubbles. Bakshy \textit{et al.} \cite{bakshy2015exposure} examines the extent of ideological diversity in social media users' online news consumption,
Bozdag \textit{et al.} \cite{bozdag_bias_2013} explores the ethical implications of algorithmic bias in personalization, and finally, Nguyen \cite{nguyen_echo_2020} explores the distinction between epistemic bubbles and echo chambers and their implications on post-truth and fake news.

\subsection{Diverse Recommendation}
Diversified recommendation aims to improve the recommendation list's diversity besides the main goal of recommendation accuracy. This is usually seen as a way to counterbalance filter bubbles since the latter are characterized by low item diversity. 
An early but significant work is by Zhou \textit{et al.} \cite{zhou_solving_2010} who define an algorithm based on heat conduction to increase diversity, effectively resolving the trade-off between diversity and accuracy. Some works propose post-processing methods that rely on a backbone recommendation model. MRR~\cite{Carbonell_Goldstein_1998} proposed a greedy method that can re-rank the output of any recommendation model, serving as a plug-in strategy to improve recommendation diversity.
Qin~\textit{et al.}\cite{Qin_Zhu} use regularization terms based on backbone recommendation to ensure the entropy. DPP~\cite{chen_fast_2018} is a popular method that uses a greedy strategy for accelerating the determinantal point process, whereas ~\cite{gillenwater2019tree} and \cite{warlop2019tensorized} further accelerate it via tree-search or factorization techniques. Li \textit{et al.} \cite{li_latent_2020} enhance diversity by increasing unexpectedness in the latent space.
In contrast, DGCN, proposed by Zheng~\textit{et al.}~\cite{zheng_dgcn_2021} was the first to design a diversified recommendation strategy for graph neural network-based models. More recently Yang \textit{et al.} \cite{yang_dgrec_2023} uses a concoction of techniques such as submodular neighbor selection, layer attention, and long-tail loss reweighting to outperform the state-of-the-art in diverse recommendation for GNN.


%% file: 5.conclusion.tex

\section{Discussion}\label{sec::discussion}
In the above exploratory study of the deep filter bubble on a popular short-video platform, we undergo a comprehensive study of the presence of variations in category coverage over the set of users based on time and category level. 
First, we define a metric and criterion that defines whether a user is "in" or "out" of a filter bubble based on whether they have been exposed to less than the median number of categories per level. Using this criterion, we discover trends in the evolution of the three-level filter bubble: the top level bubble starts out narrow and expands over time, whereas the middle level narrows and the deepest level remains mostly invariant. We also depict the probability distribution of coverages for the three levels, showing that despite having a lower relative filter bubble, deeper layers are prone to having lower coverage overall. This is due to the tree structure of the data which naturally causes narrowing of diversity per level as the categories get deeper. 
After qualifying the evolution of the filter bubble at each later over time, we analyze three correlating factors that could be partially responsible for the formation of the deep filter bubble: specific categories, user characteristics, and feedback type. By analyzing changes in dominant categories, we find that users may be led into a bubble based on top-layer categories identified at the beginning of their usage. Concerning user characteristics, we find that age and gender pose the largest risk of deep filter bubbles. 
Finally, we note that implicit feedback can serve as a mediator to the filter bubble, whereas explicit feedback can exacerbate it. 

\section{Conclusion and Future Work}\label{sec::conclusion}
In light of the above findings, we suggest some ways that recommender systems can be designed to alleviate the deep filter bubble problem. First of all, the recommender models should not over-value preferences learned early, and instead maintain top-level diversity over time so that the user is not caught in a deep bubble due to narrow early top-level exposure. 
Secondly, concerning the demographic findings, we suggest that platforms introduce measures to prevent users of certain groups from being more susceptible to the deep filter bubble, namely, younger people, females, and those from lower-tier cities. 
Finally, we suggest that recommenders should bestow an appropriate amount of importance on various forms of feedback. For example, while explicit feedback usually signifies stronger interest, implicit feedback can also be used to learn diverse user preferences and alleviate the deep filter bubble. 

For future work, we propose to delve deeper into the effect of the partial exposure of each level's categories on the deep filter bubble. We also propose to delve into the dynamics between user-platform interactions to understand how different types of behaviors could result in different treatment by the recommender.